\documentclass[10pt,twocolumn,letterpaper]{article}

\usepackage{cvpr}
\usepackage{times}
\usepackage{epsfig}
\usepackage{graphicx}
\usepackage{amsmath}
\usepackage{amssymb}
\usepackage{multirow}
\usepackage{color}
\usepackage{caption}
\usepackage{booktabs}
\usepackage{subcaption}
\usepackage{cite}
\usepackage{enumitem}

\usepackage{mmstyle}

\usepackage[pagebackref=true,breaklinks=true,letterpaper=true,colorlinks,bookmarks=false]{hyperref}

\cvprfinalcopy 

\def\cvprPaperID{1142} 

\ifcvprfinal\fi

\begin{document}

\title{Detecting Visual Relationships with Deep Relational Networks}

\author{Bo Dai\qquad\qquad~Yuqi Zhang\qquad\qquad~Dahua Lin\\
Department of Information Engineering, The Chinese University of Hong Kong\\
{\footnotesize\texttt{db014@ie.cuhk.edu.hk}~~~~~~\texttt{zy016@ie.cuhk.edu.hk}~~~~~~\texttt{dhlin@ie.cuhk.edu.hk}}
}

\maketitle
\begin{abstract}
Relationships among objects play a crucial role in image understanding.
Despite the great success of deep learning techniques in recognizing 
individual objects, reasoning about the relationships among objects
remains a challenging task.
Previous methods often treat this as a classification problem,
considering each type of relationship (e.g.~``ride'') or each distinct
visual phrase (e.g.~``person-ride-horse'') as a category.
Such approaches are faced with significant difficulties caused by
the high diversity of visual appearance for each kind of relationships
or the large number of distinct visual phrases. 
We propose an integrated framework to tackle this problem.
At the heart of this framework is the Deep Relational Network,
a novel formulation designed specifically for exploiting the 
statistical dependencies between objects and their relationships. 
On two large data sets, the proposed method achieves substantial 
improvement over state-of-the-art. 
\end{abstract}
\section{Introduction}
\label{sec:intro}
\begin{figure}[t]
\centering
\includegraphics[width=0.5\textwidth]{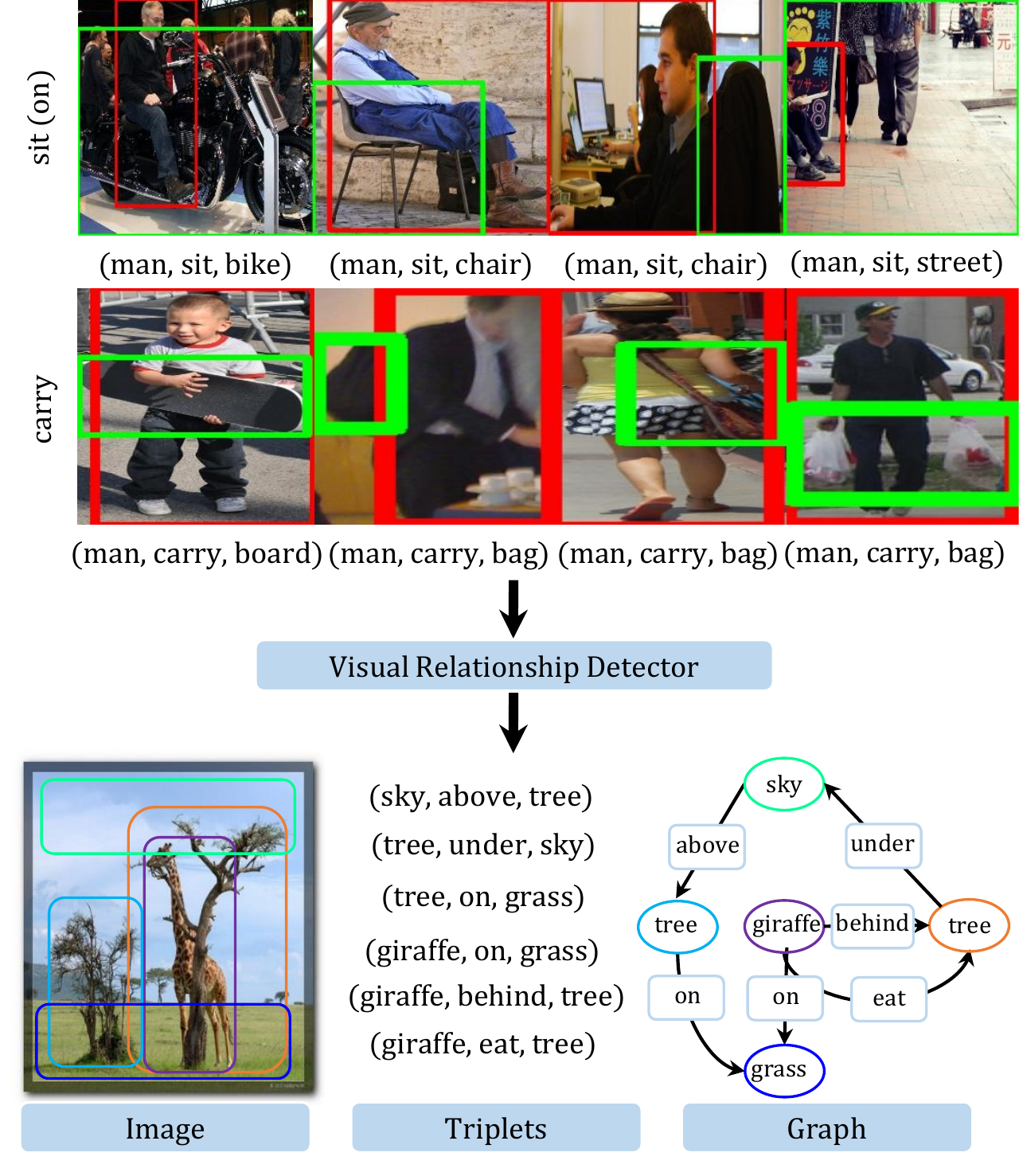}
\caption{\small 
Visual relationships widely exist in real-world images.
Here are some examples from the VRD~\cite{lu2016visual} dataset,
with relationship predicates \emph{``sit''} and \emph{``carry''}.
We develop a method that can effectively detect such 
relationships from a given image.
On top of that, a scene graph can be constructed.}
\label{fig:teaser}
\end{figure}

Images in the real world often involve multiple objects that interact
with each other. 
To understand such images, being able to recognize individual objects
is generally not sufficient. 
The \emph{relationships} among them also contain crucial messages.
For example, image captioning, a popular application in computer vision,
can generate richer captions based on relationships in addition to objects in the images.
Thanks to the advances in deep learning, the past several years witness
remarkable progress in several key tasks in computer vision, such as
\emph{object recognition}~\cite{ren2015fasterrcnn}, 
\emph{scene classification}~\cite{zhou2016places}, and 
\emph{attribute detection}~\cite{zhang2014panda}.
However, visual relationship detection remains a \emph{very difficult} task. 
On Visual Genome~\cite{krishnavisualgenome}, a large dataset designed for 
structural image understanding, the state-of-the-art
can only obtain $11.79\%$ of Recall@50~\cite{lu2016visual}.
This performance is clearly far from being satisfactory.
 
A natural approach to this problem is to treat it as a classification task.
Early attempts~\cite{sadeghi2011recognition} used to consider different
combinations of objects and relationship predicates (known as \emph{visual phrases}) 
as different classes. 
While it may work in a restricted context where the number
of possible combinations is moderate, 
such strategy would be met with a fundamental difficulty in general
-- an extremely large number of imbalanced classes.
As a case in point, Visual Genome~\cite{krishnavisualgenome} contains 
over $75K$ distinct visual phrases, and the number of samples for each phrase 
ranges from just a handful to over $10K$. 
Even the most sophisticated classifier would suffer
facing such a large and highly imbalanced class space.

An alternative strategy is to consider each type of relationship predicates
as a class. Whereas the number of classes is drastically smaller,
along with this change also comes with an undesirable 
implication, namely the substantially increased diversity within each class.
To be more specific, phrases with different object categories
are considered to be in the same class, as long as they have 
the same type of relationship predicates.
Consequently, the images in each class are highly diverse --
some images in the same class may even share nothing in common,
\eg~\emph{``mountain-near-river''} and \emph{``person-near-dog''}.
See Figure \ref{fig:teaser} for an illustration.
Our experiments suggest that even with the model capacity of deep networks,
handling the intra-class diversity at this level remains very difficult.

In this work, we develop a new framework to tackle the problem 
of \emph{visual relationship detection}.
This framework formulates the prediction output as a triplet
in the form of \emph{(subject, predicate, object)}, and
jointly infers their class labels by exploiting 
two kinds of relations among them, namely
\emph{spatial configuration} and \emph{statistical dependency}.
Such relations are ubiquitous, informative, and more importantly
they are often more reliable than visual appearance.

It is worth emphasizing that the formulation of the proposed model
is significantly different from previous relational models such as
conditional random fields (CRFs) \cite{lafferty2001conditional}. 
Particularly, in our formulation, the statistical inference procedure
is embedded into a deep neural network 
called \emph{Deep Relational Network (DR-Net)} via iteration unrolling.
The formulation of DR-Net moves beyond the conventional scope, extending
the expressive power of Deep Neural Networks (DNNs) to relational modeling.
This new way of formulation also allows the model parameters to be learned 
in a discriminative fashion, using the latest techniques in deep learning.  
On two large datasets, the proposed framework outperforms not only 
the classification-based methods but also the CRFs based on deep potentials.

To sum up, the major contributions of this work consist in two aspects:
(1) DR-Net, a novel formulation that combines the strengths of statistical
models and deep learning; and
(2) an effective framework for visual relationship detection\footnote{code available at \url{github.com/doubledaibo/drnet}}, 
which brings the state-of-the-art to a new level.

\section{Related Work}
\label{sec:relwork}

Over the past decade, there have been a number of studies that explore
the use of \emph{visual relationships}.
Earlier efforts often focus on \emph{specific} types of relationships, 
such as positional relations~\cite{gupta2008beyond, johnson2015image, galleguillos2008object, choi2013understanding, kulkarni2011baby, elliott2013image}
and actions (\ie~interactions between objects)~\cite{yao2010grouplet, gkioxari2015contextual, regneri2013grounding, thomason2014integrating, ramanathan2015learning, rohrbach2013translating, guadarrama2013youtube2text, antol2014zero, elhoseiny2015sherlock, farhadi2010every,xiong2015recognize}.
In most of these studies, relationships are usually extracted using simple 
heuristics or hand-crafted features, and used as 
an auxiliary components to facilitate other tasks, such as
object recognition~\cite{galleguillos2010context, sivic2005discovering, kumar2010efficiently, choi2010exploiting, ladicky2010graph, salakhutdinov2011learning, rabinovich2007objects, fidler2007towards, russell2006using},
image classification and retrieval~\cite{mensink2014costa, gong2014multi},
scene understanding and generation~\cite{zitnick2013learning, hoiem2008putting, chang2014semantic, yao2012describing, izadinia2014incorporating, gould2008multi, berg2012understanding},
as well as text grounding\cite{plummer2015flickr30k, karpathy2014deep, rohrbach2015grounding}.
They are essentially different from our work, which aims to provide a method
dedicated to \emph{generic} visual relationship detection.
On a unified framework, our method can recognize a wide variety of relationships, 
such as relative positions (\emph{``behind''}), actions (\emph{``eat''}), 
functionals (\emph{``part of''}), and comparisons (\emph{``taller than''}).

Recent years have seen new methods developed specifically for detecting visual relationships. 
An important family of methods~\cite{das2013thousand, divvala2014learning, sadeghi2011recognition}
consider each distinct combination of object categories and relationship predicates as
a distinct class (often referred to as a \emph{visual phrase}). 
Such methods would face difficulties in a general context, 
where the number of such combinations can be very large. 
An alternative paradigm that considers relationship predicates and object categories separately
becomes more popular in recent efforts. 
Vedantam \etal~\cite{vedantam2015learning} presented a study along this line
using synthetic clip-arts. This work, however, relies on multiple synthetic attributes
that are difficult to obtain from natural images. 
Fang \etal~\cite{fang2015captions} proposed to incorporate relationships in an image captioning
framework.
This work treats object categories and relationship predicates uniformly as words,
and does not discuss how to tackle the various challenges in relationship detection. 

The method proposed recently by Lu \etal~\cite{lu2016visual} is the most related. 
In this method, pairs of detected objects are fed to a classifier, which combines
appearance features and a language prior for relationship recognition. 
Our method differs in two aspects:
(1) We exploit both spatial configurations and statistical dependencies among 
\emph{relationship predicates}, \emph{subjects}, and \emph{objects},
via a Deep Relational Network, instead of simply fusing them as different features.
(2) Our framework, from representation learning to relational modeling, is integrated 
into a single network that is learned in an end-to-end fashion.
Experiments show that the proposed framework performs substantially better in 
all different task settings. For example, on two large datasets, 
the \emph{Recall@50} of relationship predicate recognition are respectively raised from 
$47.9\%$ to $80.8\%$ and from $53.5\%$ to $88.3\%$.



\section{Visual Relationship Detection}
\label{sec:ovalfrm}

\begin{figure*}
    \centering
    \includegraphics[width=\textwidth]{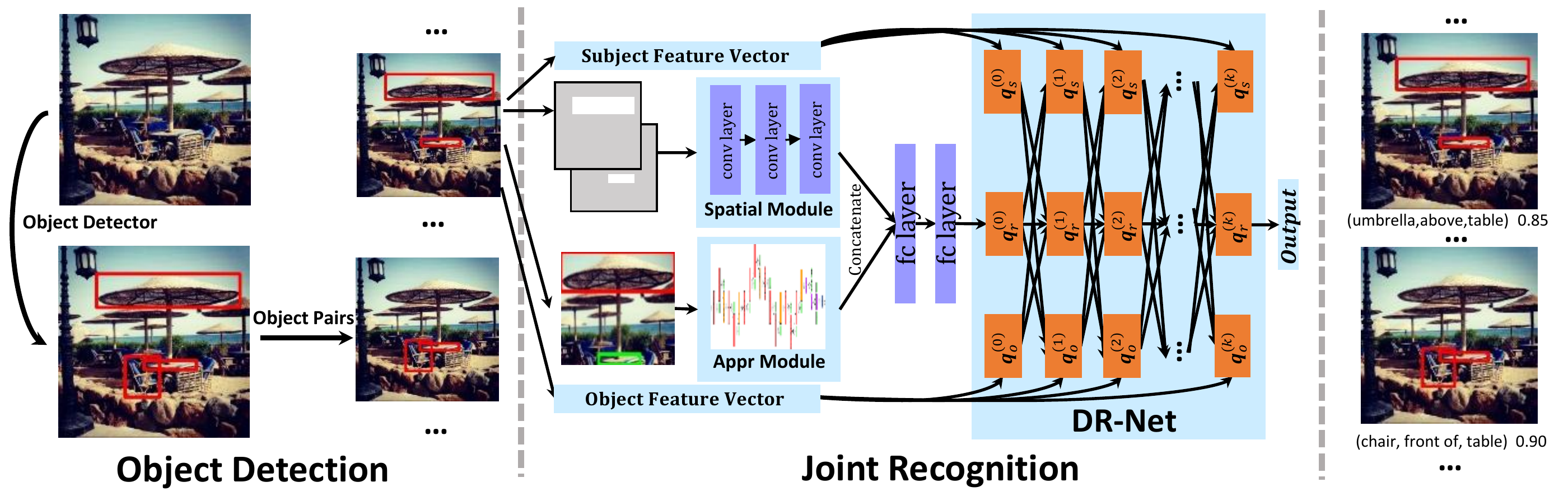}
    \caption{\small
    The proposed framework for visual relationship detection.
    Given an image, it first employs an object detector to locate individual objects.
    Each object also comes with an appearance feature.
    For each pair of objects, the corresponding local regions and the spatial masks 
    will be extracted, which, together with the appearance features 
    of individual objects, will be fed to the DR-Net. 
    The DR-Net will jointly analyze all aspects
    and output $\vq_s$, $\vq_r$, and $\vq_o$, the predicted category 
    probabilities for each component of the triplet. 
    Finally, the triplet $(s, r, o)$ will be derived by choosing the most probable 
    categories for each component.}
    \label{fig:overview}
\end{figure*}

Visual relationships play a crucial role in image understanding.
Whereas a relationship may involve multiple parties in general,
many important relationships, including
\emph{relative positions} (\eg~\emph{``above''}) and
\emph{actions} (\eg~\emph{``ride''}) occur between exactly two objects.  
In this paper, we focus on such relationships.
In particular, we follow a widely adopted convention~\cite{sadeghi2011recognition,lu2016visual}
and characterize each visual relationship by a \emph{triplet}
in the form of $(s, r, o)$, 
\eg~\emph{(girl, on, horse)} and \emph{(man, eat, apple)}.
Here, $s$, $r$, and $o$ respectively
denote the \emph{subject category}, the \emph{relationship predicate}, and
the \emph{object category}. 
The task is to locate all visual relationships from a given image, 
and infer the triplets. 

\subsection{Overall Pipeline}

As mentioned, there are two different paradigms for relationship detection:
one is to consider each distinct triplet as a different category 
(also known as \emph{visual phrases}~\cite{sadeghi2011recognition}), 
the other is to recognize each component individually.
The former is not particularly suitable for generic applications,   
due to difficulties like the excessively large number of classes and 
the imbalance among them.
In this work, we adopt the latter paradigm and aim to take its 
performance to a next level.
Particularly, we focus on developing a new method that can effectively 
capture the rich relations (both \emph{spatial} and \emph{semantic}) 
among the three components in a triplet 
and exploit them to improve the prediction accuracy.

As shown in Figure~\ref{fig:overview}, 
the overall pipeline of our framework comprises three stages, 
as described below.

\textbf{(1) Object detection.} 
Given an image, we use an object detector to locate a set of candidate objects.
In this work, we use Faster RCNN~\cite{ren2015fasterrcnn} for this purpose.
Each candidate object comes with a bounding box and 
an appearance feature,
which will be used in the joint recognition stage for
predicting the object category.

\textbf{(2) Pair filtering.}
The next step is to produce a set of \emph{object pairs} from the detected
objects. With $n$ detected objects, we can form $n (n - 1)$ pairs.
We found that a considerable portion of these pairs are \emph{obviously} 
meaningless and it is unlikely to recognize important relationships therefrom.
Hence, we introduce a low-cost neural network to filter out such pairs,
so as to reduce the computational cost of the next stage. 
This filter takes into account both the spatial configurations 
(\eg~objects too far away are unlikely to be related)
and object categories 
(\eg~certain objects are unlikely to form a meaningful relationship). 

\textbf{(3) Joint recognition.}
Each retained pair of objects will be fed to the \emph{joint recognition}
module. 
Taking into account multiple factors and their relations,
this module will produce a triplet as the output.

\begin{figure}
    \centering
    \includegraphics[height=0.22\textwidth]{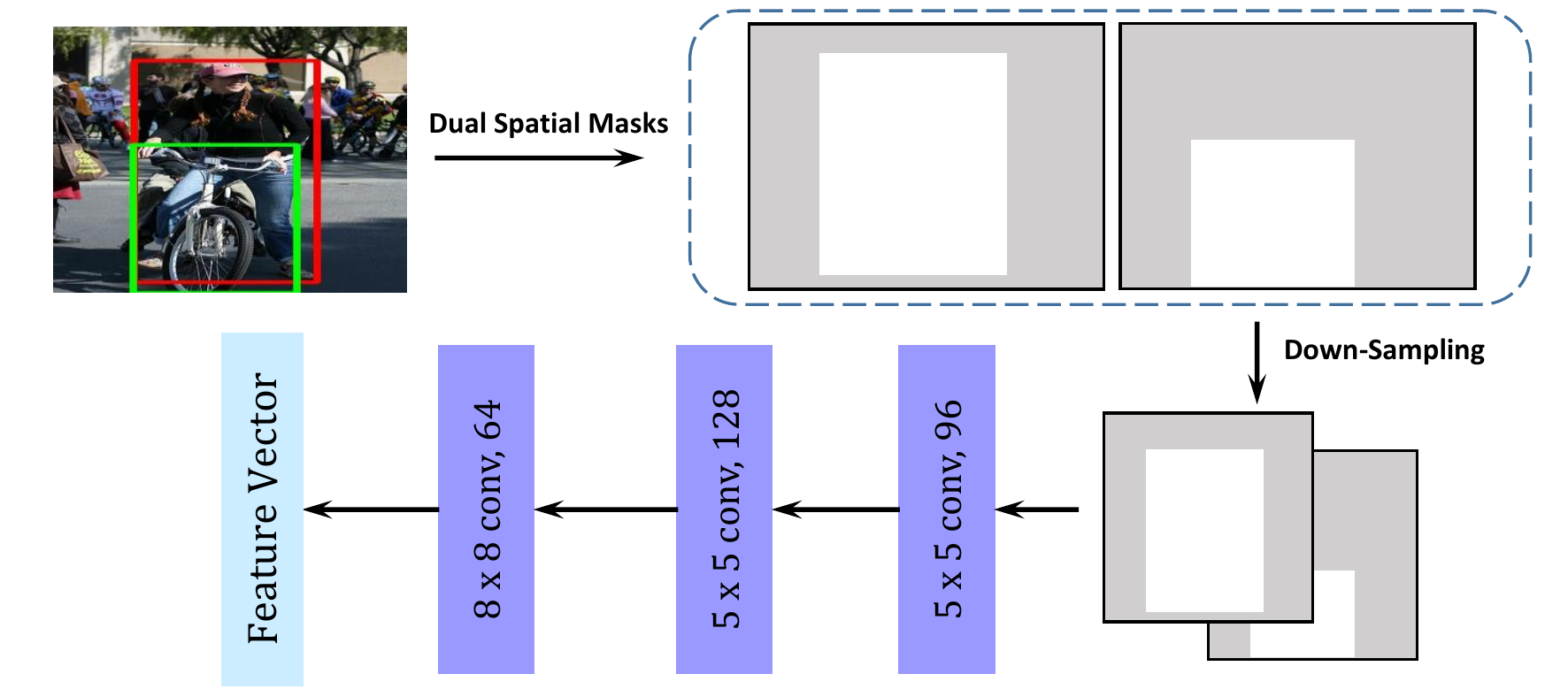}
    \caption{\small This figure illustrates the process of spatial feature vector generation. 
		The structure of our spatial module is also presented in this figure.}
    \label{fig:sp_module}
\end{figure}

\subsection{Joint Recognition}

In joint recognition, multiple factors are taken into consideration.
These factors are presented in detail below.

\textbf{(1) Appearance.} 
As mentioned, each detected object comes with an appearance feature,
which can be used to infer its category. 
In addition, the type of the relationship may also be
reflected in an image visually.
To utilize this information, we extract an appearance feature for each
\emph{candidate pair} of objects, by applying a CNN~\cite{Simonyan14c,he2015deep} 
to an \emph{enclosing box}, 
\ie~a bounding box that encompasses both objects with a small margin.
The appearance inside the enclosing box captures not only the objects themselves
but also the surrounding context, which is often useful when
reasoning about the relationships.

\textbf{(2) Spatial Configurations.} 
The relationship between two objects is also reflected
by the spatial configurations between them, 
\eg~their relative positions and relative sizes.
Such cues are complementary to the appearance of individual objects,
and resilient to photometric variations, \eg~the changes in illumination.

To leverage the spatial configurations, we are facing a question:
{\em how to represent it in a computer?}
Previous work~\cite{johnson2015image} suggests a list of geometric measurements.
While simple, this way may risk missing certain aspects of the configurations.
In this work, we instead use \emph{dual spatial masks} as the representation,
which comprise two binary masks, one for the subject and the other for the object.
The masks are derived from the bounding boxes and may overlap with each other,
as shown in Figure~\ref{fig:sp_module}.
The masks are down-sampled to the size $32 \times 32$, 
which we found empirically is a good balance between fidelity and cost.
(We have tried mask sizes of $8$, $16$, $32$, $64$ and $128$, 
resulting top-1 recalls are $0.47$, $0.48$, $0.50$, $0.51$ and $0.51$.)
The dual spatial masks for each candidate pair will be compressed into
a $64$-dimensional vector via three convolutional layers.

\textbf{(3) Statistical Relations.} 
In a triplet $(s, r, o)$, 
there exist strong statistical dependencies 
between the relationship predicate $r$ and the object categories $s$ and $o$. 
For example, \emph{(cat, eat, fish)} is common, 
while \emph{(fish, eat, cat)} or \emph{(cat, ride, fish)} is very unlikely. 
On Visual Genome, the entropy of the prior distribution $p(r)$ is $2.88$,
while that of the conditional distribution $p(r |s, o)$ is $1.21$.
This difference is a clear evidence of the statistical dependency. 

To exploit the statistical relations, we propose 
\emph{Deep Relational Network (DR-Net)},
a novel formulation that incorporates statistical relational modeling
into a deep neural network framework. 
In our experiments, we found that the use of such relations can effectively
resolve the ambiguities caused by visual or spatial cues, thus
substantially improving the prediction accuracy.

\textbf{(4) Integrated Prediction.} 
Next, we describe how these factors are actually combined.
As shown in Figure~\ref{fig:overview},
for each candidate pair, the framework extracts the appearance 
feature and the spatial feature, respectively via the 
appearance module and the spatial module. 
These two features are subsequently concatenated and further compressed
via two fully-connected layers.
This \emph{compressed pair feature}, together with the appearance features
of individual objects will be fed to the DR-Net for joint inference.
Through multiple inference units, whose parameters capture the 
statistical relations among triplet components, 
the \emph{DR-Net} will output the posterior probabilities of $s$, $r$, and $o$.
Finally, the framework produces the prediction by choosing
the most probable classes for each of these components. 

In the training, all stages in our framework, 
namely object detection, pair filtering and joint recognition are trained respectively.
As for joint recognition,
different factors will be integrated into a single network and jointly fine-tuned to 
maximize the joint probability of the ground-truth triplets.

\section{Deep Relational Network}
\label{sec:drnet}

As shown above, there exist strong statistical relations among 
the object categories $s$ and $o$ and the relationship predicates $r$. 
Hence, to accurately recognize visual relationships, it is important
to exploit such information, especially when the visual cues are ambiguous.
 

\subsection{Revisit of CRF}
\label{sec:crf}

The \emph{Conditional Random Field (CRF)}~\cite{lafferty2001conditional} 
is a classical formulation to incorporate statistical relations into a discriminative task.
Specifically, for the task of recognizing visual relationships, the CRF
can be formulated as
{\small
\begin{equation} \label{eq:crf0}
	p(r, s, o | \vx_r, \vx_s, \vx_o)
	= \frac{1}{Z} \exp \left( \Phi(r, s, o | \vx_r, \vx_s, \vx_o; \mW) \right).
\end{equation}
}
Here, $\vx_r$ is the \emph{compressed pair feature} that combines both the 
appearance of the enclosing box and the spatial configurations;
$\vx_s$ and $\vx_o$ are the appearance features respectively for the subject and the object;
$\mW$ denotes the model parameters;
and $Z$ is the normalizing constant, whose value depends on the parameters $\mW$.
The joint potential $\Phi$ can be expressed as a sum of individual potentials as
{\small
\begin{align} \label{eq:crf1}
	\Phi 
	&= \psi_a(s | \vx_s; \mW_a) 
	 + \psi_a(o | \vx_o; \mW_a) 
	 + \psi_r(r | \vx_r; \mW_r) \notag \\
	&+ \varphi_{rs}(r, s | \mW_{rs})
	 + \varphi_{ro}(r, o | \mW_{ro})
	 + \varphi_{so}(s, o | \mW_{so}).
\end{align}
}
Here, the unary potential $\psi_a$ associates individual objects with their appearance;
$\psi_r$ associates the relationship predicate with the feature $\vx_r$;
while the binary potentials $\varphi_{rs}$, $\varphi_{ro}$ and $\varphi_{so}$ capture 
the statistical relations among the relationship predicate $r$,  
the subject category $s$, and the object category $o$.

CRF formulations like this have seen wide adoption in computer vision literatures~\cite{zheng2015conditional, quattoni2004conditional}
over the past decade, and have been shown to be a viable way to capture statistical dependencies.
However, the success of CRF is limited by several issues:
First, learning CRF requires computing the normalizing constant $Z$, 
which can be very expensive and even intractable, especially when 
cycles exist in the underlying graph, like the formulation above. 
Hence, approximations are often used to circumvent this problem, 
but they sometimes result in poor estimates.
Second, when cyclic dependencies are present, 
variational inference schemes 
such as mean-field methods~\cite{koltun2011efficient} and loopy belief propagation~\cite{pearl1988probabilistic},
are widely used to simplify the computation.
This often leaves a gap between the objective of inference and
that of training, thus leading to suboptimal results.

\subsection{From CRF to DR-Net}


Inspired by the success of deep neural networks~\cite{he2015deep, Simonyan14c}, 
we explore an alternative approach to relational modeling, 
that is, to \emph{unroll} the inference into a feed-forward network.

Consider the CRF formulated above. 
Given $s$ and $o$, then the posterior distribution of $r$ is given by
\begin{multline}
	p(r | s, o, \vx_r; \mW) \propto \exp \left(
		\psi_r(r | \vx_r; \mW_r) + \right. \\
		\left. \varphi_{rs}(r, s | \mW_{rs}) 
		+ \varphi_{ro}(r, o | \mW_{ro})
	\right).
\end{multline}
In typical formulations, 
$\psi_r(r | \vx_r)$ is often devised to be a linear functional of $\vx_r$ for each $r$.
Let $\mW_{rs}$ and $\mW_{ro}$ be matrices such that
$\mW_{rs}(r, s) = \varphi_{rs}(r, s)$ and 
$\mW_{ro}(r, o) = \varphi_{ro}(r, o)$, and
let $\vq_r$ be a vector of the posterior probabilities for $r$, then
the formula above can be rewritten 
as\footnote{A proof of this statement is provided in the supplemental materials.}
\begin{equation} \label{eq:rsol0}
	\vq_r = \vsigma \left(
		\mW_r \vx_r + \mW_{rs} \vone_s + \mW_{ro} \vone_o
	\right).	
\end{equation}
Here, $\vsigma$ denotes the \emph{softmax} function.
$\vone_s$ and $\vone_o$ are one-hot indicator vectors for $s$ and $o$.
It can be shown that this is the optima to the optimization problem below:
\begin{multline}
	\max_\vq \ E_q \left[ \psi_r(r | \vx_r; \mW_r) + \right. \\
	\left. \varphi_{rs}(r, s | \mW_{rs}) + \varphi_{ro}(r, o | \mW_{ro}) \right] + H_q(\vq).
\end{multline}
Based on this optimization problem, the solution given in Eq.\eqref{eq:rsol0} can be 
generalized to the case where $s$ and $o$ are not deterministic and 
the knowledge of them are instead given by probabilistic vectors $\vq_s$ and $\vq_o$, as follows:
\begin{equation} \label{eq:rso}
	\vq_r = \vsigma \left(
		\mW_r \vx_r + \mW_{rs} \vq_s + \mW_{ro} \vq_o
	\right).	
\end{equation}
Similar derivation also applies to the inference of $s$ and $o$ conditioned on other components. 
Together, we can obtain a set of \emph{updating formulas} as below:
\begin{align} \label{eq:runit}
        \vq'_s & = \vsigma \left( \mW_a \vx_s + \mW_{sr} \vq_r + \mW_{so} \vq_o \right), \notag \\
        \vq'_r & = \vsigma \left( \mW_r \vx_r + \mW_{rs} \vq_s + \mW_{ro} \vq_o \right), \notag \\
        \vq'_o & = \vsigma \left( \mW_a \vx_o + \mW_{os} \vq_s + \mW_{or} \vq_r \right).
\end{align}
These formulas take the current probability vectors $\vq_s$, $\vq_r$, and $\vq_o$ as inputs, and
output the updated versions $\vq'_s$, $\vq'_r$ and $\vq'_o$. 
From the perspective of neural networks, these formulas can also be viewed as a \emph{computing layer}. 
In this sense, the iterative updating procedure can be \emph{unrolled} into a network that comprises 
a sequence of such layers. 
We call this network the \emph{Deep Relational Network (DR-Net)}, as it relates multiple variables, and
refer to its building blocks, \ie~the computing layers mentioned above, as \emph{inference units}.

\paragraph{Discussion}

DR-Net is for relational modeling,
which is different from those methods for feature/modality combination.
Specifically, \emph{object categories} and \emph{relationship predicates} are
two distinct domains that are statistically related. 
The former is not an extra feature of the latter; while the latter is 
not a feature of the former either. 
DR-Net captures the relations between them via the links in the 
inference units, rather than combining them using a fusion layer. 

The basic formulation in Eq.\ref{eq:runit} comes with several symmetry constraints: 
$\mW_{sr} = \mW_{rs}^T$, $\mW_{so} = \mW_{os}^T$, and $\mW_{ro} = \mW_{or}^T$. 
In addition, all inference units share the same set of weights. 
However, from a pragmatic standpoint, one may also consider lifting these constraints,
\eg~allowing each inference units to have their own weights. 
This may potentially increase the expressive power of the network.
We will compare these two settings, namely with and without weight sharing, in our experiments. 

A DR-Net can also be considered as a special form of the Recurrent Neural Network (RNN)
-- at each step it takes in a fixed set of inputs, 
\ie~the observed features $\vx_s$, $\vx_r$, and $\vx_o$, and
refines the estimates of posterior probabilities.

\subsection{Comparison with Other Formulations}


There are previous efforts that also explore the incorporation of relational structures 
with deep networks~\cite{chen2015learning, zheng2015conditional, schwing2015fully, belanger2015structured}. 
The deep structured models presented in \cite{chen2015learning,schwing2015fully,wu2016deep} combine
a deep network with an MRF or CRF on top to capture the relational structures among their outputs. 
In these works, classical message-passing methods are used in training and inference.
Zheng \etal~\cite{zheng2015conditional} proposed a framework for image segmentation,
which adopts an apparently similar idea, that is, 
to reformulate a structured model into a neural network by turning inference updates into neural layers. 
In addition to the fact that this work is in a fundamentally different domain 
(high-level understanding vs. low-level vision), 
they focused on capturing dependencies among elements in the same domain,
\eg~those among pixel-wise labels.
From a technical view, DR-Net is more flexible, 
\eg~it can handle graphs with nodes of different cardinalities and edges of different types.
In~\cite{zheng2015conditional}, the message passing among pixels is \emph{approximately} 
instantiated using CNN filters and this is primarily suited for grid structures; 
while in DR-Net, the inference steps are exactly reproduced using fully-connected layers.
Hence, it can be applied to capture relationships of arbitrary structures.
SPENs introduced in \cite{belanger2015structured} define a neural network 
serving as an energy function over observed features for multi-label classification.
SPENs are used to measure the consistency of configurations, 
while DR-Net is used to find a good configuration of variables.
Also, no inference unrolling is involved in SPENs learning.


\section{Experiments}
\label{sec:exprt}

\begin{table*}[t]
    \centering
    \small
\begin{tabular}{c|c|c|c|c|c|c|c}
        &  & \multicolumn{2}{c}{Predicate Recognition} & \multicolumn{2}{|c}{Union Box Detection} & \multicolumn{2}{|c}{Two Boxes Detection} \\
        \cline{3-8}
        & & Recall@50 & Recall@100 & Recall@50 & Recall@100 & Recall@50 & Recall@100 \\
	\hline
        \multirow{5}{*}{\rotatebox{90}{\textbf{VRD}}}  
        & VP \cite{sadeghi2011recognition}    & 0.97 & 1.91 & 0.04 & 0.07 & - & - \\
        & Joint-CNN\cite{fang2015captions}    & 1.47 & 2.03 & 0.07 & 0.09 & 0.07 & 0.09 \\
        & VR \cite{lu2016visual}              & 47.87 & 47.87 & 16.17 & 17.03 & 13.86 & 14.70 \\
        & DR-Net	       		      & \textbf{80.78} & \textbf{81.90} & 19.02 & 22.85 & 16.94 & 20.20 \\
	& DR-Net + pair filter                            & - & - & \textbf{19.93} & \textbf{23.45} & \textbf{17.73} & \textbf{20.88} \\
        \hline
        \hline
        \multirow{5}{*}{\rotatebox{90}{\textbf{sVG}}}  
        & VP \cite{sadeghi2011recognition}    & 0.63 & 0.87 & 0.01 & 0.01 & - & - \\
        & Joint-CNN\cite{fang2015captions}    & 3.06 & 3.99 & 1.24 & 1.60 & 1.21 & 1.58  \\
        & VR \cite{lu2016visual}              & 53.49 & 54.05 & 13.80 & 17.39 & 11.79 & 14.84  \\
        & DR-Net                             & \textbf{88.26} & \textbf{91.26} & 20.28 & 25.74 & 17.51 & 22.23 \\
	& DR-Net + pair filter			      & - & - & \textbf{23.95} & \textbf{27.57} & \textbf{20.79} & \textbf{23.76} 
\end{tabular}
    \caption{\small Comparison with baseline methods, using \emph{Recall@50} and \emph{Recall@100} as the metrics.
    We use ``-'' to indicate \emph{``not applicable''}. 
    For example, no results are reported for \emph{DR-Net + pair filter} on Predicate Recognition, as in this setting,
    pairs are given, and thus pair filtering can not be applied.
    Also, no results are reported for \emph{VP} on Two Boxes detection, as VP detects the entire instance as 
    a single entity.    
    }
    \label{tab:itr_rst}
\end{table*}

\begin{table*}[t]
    \centering
    \small
\begin{tabular}{c|c|c|c|c|c|c|c|c|c}
	&  & A$_1$ & A$_2$ & S & A$_1$S & A$_1$SC & A$_1$SD & A$_2$SD &A$_2$SDF \\
	\hline
	\multirow{3}{*}{\rotatebox{90}{\textbf{VRD}}}
	& Predicate Recognition & 63.39 & 65.93 & 64.72 & 71.81 & 72.77 & 80.66 & \textbf{80.78} & -\\
	& Union Box Detection & 12.01 & 12.56 & 13.76 & 16.04 & 16.37 & 18.15 & \textbf{19.02} & \textbf{19.93} \\
	& Two Boxes Detection & 10.71 & 11.22 & 12.16 & 14.38 & 14.66 & 16.12 & \textbf{16.94} & \textbf{17.73} \\
	\hline
	\hline
	\multirow{3}{*}{\rotatebox{90}{\textbf{sVG}}}
	& Predicate Recognition & 72.13 & 72.54 & 75.18 & 79.10 & 79.18 & 88.00 & \textbf{88.26} & -\\
	& Union Box Detection & 13.24 & 13.84 & 14.01 & 16.04 & 16.08 & 20.21 & \textbf{20.28} & \textbf{23.95} \\
	& Two Boxes Detection & 11.35 & 11.98 & 12.07 & 13.77 & 13.81 & 17.42 & \textbf{17.51} & \textbf{20.79}
\end{tabular}
    \caption{\small Comparison of different variants of the proposed method, using \emph{Recall@50} as the metric.}
    \label{tab:cfg_rst}
\end{table*}

\begin{table*}
    \centering
   \small
    \setlength{\tabcolsep}{1pt}
    \begin{tabular}{c|c|c|c|c}
         & \includegraphics[width=0.2\textwidth,height=0.2\textwidth]
        {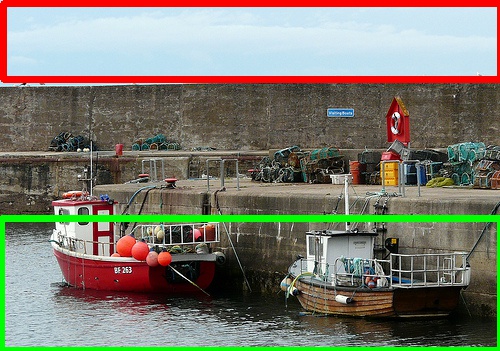} 
                    & \includegraphics[width=0.2\textwidth,height=0.2\textwidth]
        {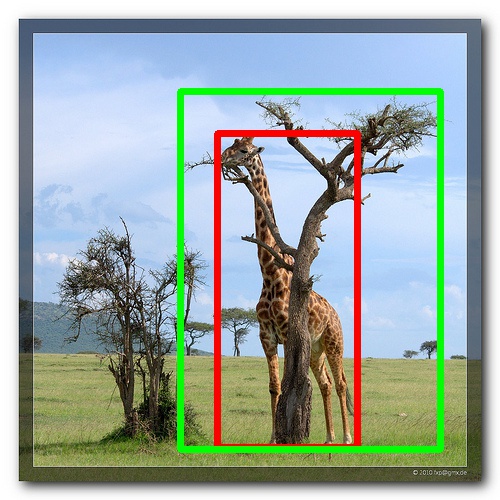}
                    & \includegraphics[width=0.2\textwidth,height=0.2\textwidth]
        {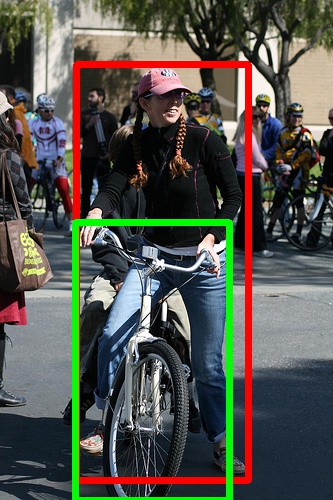}
                    & \includegraphics[width=0.2\textwidth,height=0.2\textwidth]
        {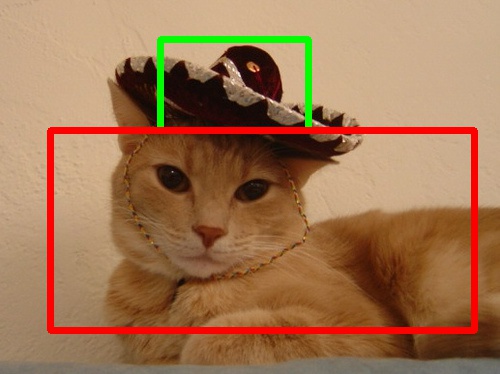}
        \\
        \hline
        VR\cite{lu2016visual} & (sky, \textbf{in}, water)
                            & (giraffe, \textbf{have}, tree)
                            & (woman, \textcolor{red}{ride}, bicycle)
                            & (cat, \textbf{have}, hat)
        \\
        A$_1$ & (sky, \textbf{on}, water)
        & (giraffe, \textbf{have}, tree) 
        & (woman, \textbf{behind}, bicycle)
        & (cat, \textbf{on}, hat)
        \\
        S & (sky, \textcolor{red}{above}, water)
        & (giraffe, \textbf{in}, tree) 
        & (woman, \textbf{wear}, bicycle)
        & (cat, \textbf{have}, hat)
        \\
        A$_1$S & (sky, \textcolor{red}{above}, water)
        & (giraffe, \textcolor{red}{behind}, tree)
        & (woman, \textbf{wear}, bicycle)
        & (cat, \textbf{have}, hat)
        \\
        A$_1$SC & (sky, \textcolor{red}{above}, water)
        & (giraffe, \textcolor{red}{behind}, tree)
        & (woman, \textcolor{red}{ride}, bicycle)
        & (cat, \textbf{have}, hat)
        \\
        A$_1$SD & (sky, \textcolor{red}{above}, water)
        & (giraffe, \textcolor{red}{behind}, tree)
        & (woman, \textcolor{red}{ride}, bicycle)
        & (cat, \textcolor{red}{wear}, hat)
    \end{tabular}
    \caption{\small This table lists predicate recognition results for some object pairs. 
    Images containing these pairs are listed in the first row,
    where the red and green boxes respectively correspond to the subjects and the objects.
    The most probable predicate predicted by different methods are listed in the following rows,
    in which \textbf{black} indicates wrong prediction and \textcolor{red}{red} indicates correct prediction.}
    \label{tab:itr_sample}
\end{table*}

We tested our model on two datasets:
(1) \textbf{VRD}: the dataset used in~\cite{lu2016visual},
containing $5,000$ images and
$37,993$ visual relationship instances that belong to
$6,672$ triplet types.
We follow the train/test split in~\cite{lu2016visual}.
(2) \textbf{sVG}: a substantially larger subset
constructed from Visual Genome~\cite{krishnavisualgenome}.
\emph{sVG} contains $108K$ images and $998K$ relationship instances
that belong to $74,361$ triplet types.
All instances are randomly partitioned into disjoint training and testing sets,
which respectively contain $799K$ and $199K$ instances.

\subsection{Experiment Settings}

\paragraph{Model training.}
In all experiments, we trained our model using Caffe\cite{jia2014caffe}.
The appearance module is initialized with a model pre-trained on ImageNet,
while the spatial module and the DR-Net are initialized randomly.
After initialization, the entire network is jointly optimized using SGD.

\paragraph{Performance metrics.}
Following~\cite{lu2016visual}, we use \emph{Recall@K} as the major performance metric,
which is the the fraction of ground-truth instances that are correctly recalled in top $K$ predictions.
Particularly, we report \emph{Recall@100} and \emph{Recall@50} in our experiments.
The reason of using \emph{recall} instead of \emph{precision} is that
the annotations are incomplete, where some true relationships might be missing.

\paragraph{Task settings.}
Like in~\cite{lu2016visual}, we studied three task settings:
\textbf{(1) Predicate recognition}: this task focuses on the accuracy of \emph{predicate} recognition,
where the labels and the locations of both the \emph{subject} and \emph{object} are given.
\textbf{(2) Union box detection}: this task treats the whole triplet as a union bounding box.
A prediction is considered correct if all three elements in a triplet $(s, r, o)$ are
correctly recognized, and the IoU between the predicted box and the ground-truth is
above $0.5$.
\textbf{(3) Two boxes detection}: this is similar to the one above,
except that it requires the IoU metrics for the subject and the object
are both above $0.5$. This is relatively more challenging.

\subsection{Comparative Results}


\paragraph{Compare with baselines.}

We compared our method with the following methods under all three task settings outlined above.
(1) \textbf{Visual Phrase(VP)}~\cite{sadeghi2011recognition}:
a representative approach that treats each distinct triplet as a different class.
and employs a DPM detector~\cite{lsvm-pami} for each class.
(2) \textbf{Joint-CNN}~\cite{fang2015captions}:
a neural network~\cite{Simonyan14c} that has $2N$+$K$-way outputs, jointly predicts the class responses
for subject, object, and relationship predicate.
(3) \textbf{Visual Relationship (VR)}~\cite{lu2016visual}:
This is the state-of-the-art and is the most closely related work.

\begin{figure}
	\centering
	\includegraphics[width=0.4\textwidth]{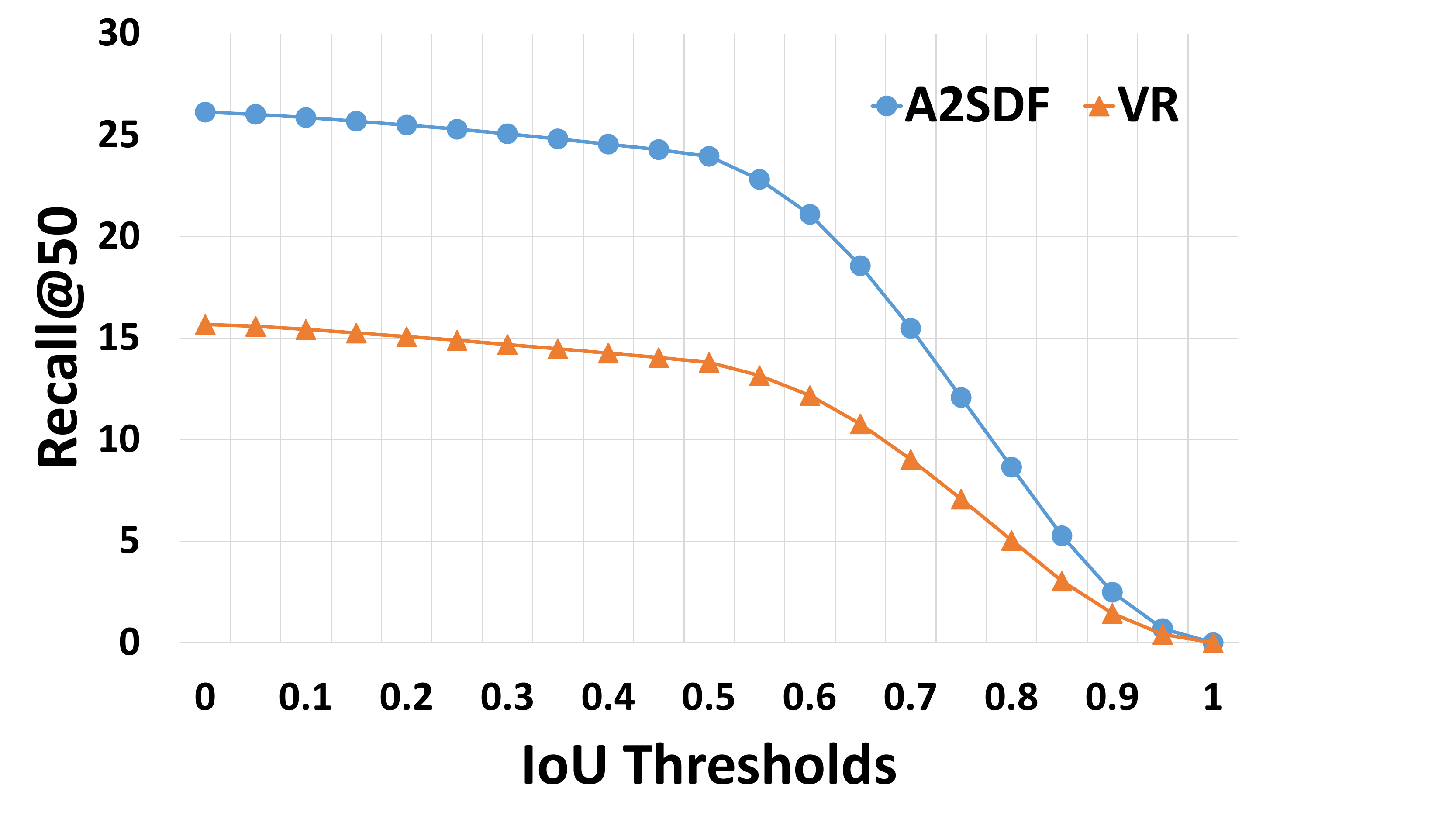}
	\caption{\small This figure shows the performance on the \emph{union-box detection} task with different IoU thresholds.}
	\label{fig:iou_relax}
\end{figure}

Table~\ref{tab:itr_rst} compares the results. On both datasets, we observed:
(1) VP~\cite{sadeghi2011recognition} performs very poorly, failing in most cases,
as it is difficult to cope with such a huge and imbalanced class space.
(2) Joint-CNN~\cite{fang2015captions} also works poorly, as
it's hard for the CNN to learn a common feature representation for both
relationship predicates and objects.
(3) VR~\cite{lu2016visual} performs substantially better than the two above.
However, the performance remains unsatisfactory.
(4) The proposed method outperforms the state-of-the-art method \emph{VR}~\cite{lu2016visual} by a considerable margin
in all three tasks.
Compared to \emph{VR}, it improves the \emph{Recall@100} of \emph{predicate recognition} by over $30\%$ on both datasets.
Thanks to the remarkably improved accuracy in recognizing the relationship predicates,
the performance gains on the other two tasks are also significant.
(5) Despite the significant gain compared to others,
the recalls on \emph{union box detection} and \emph{two boxes detection}
remains weak. This is primarily ascribed to the limitations of the object detectors.
As shown in Figure \ref{fig:iou_relax}, we observe that
the object detector can only obtain about $30\%$ of object recall, measured by \emph{Recall@50}.
To improve on these tasks, a more sophisticated object detector is needed.

\paragraph{Compare different configs.}

We also compared different variants of the proposed method, in order
to identify the contributions of individual components listed below:
(1)\textbf{Pair (F)ilter}: the pair filter discussed in section \ref{sec:ovalfrm},
used to filter out object pairs with trivial relationships.
(2)\textbf{(A)ppearance Module}: the appearance module, which has two versions,
\emph{A$_1$}: based on VGG16~\cite{Simonyan14c}, which is also the network used in VR~\cite{lu2016visual},
\emph{A$_2$}: based on ResNet101~\cite{he2015deep}.
(3)\textbf{(S)patial Module}: the network to capture the spatial configs, as mentioned in section \ref{sec:ovalfrm}.
(4)\textbf{(C)RF}: a classical CRF formulation, used as a replacement of the DR-Net to capture statistical dependencies.
(5)\textbf{(D)R-Net}: the DR-Net discussed in section \ref{sec:drnet}.
The name of a configuration is the concatenation of abbrevations of involved components,
\eg, the configuration named \emph{A$_1$SC} contains an appearance module based on VGG16, a spatial module, and a CRF.

In Table \ref{tab:cfg_rst},
we compared \emph{A$_1$}, \emph{A$_2$}, \emph{S}, \emph{A$_1$S}, \emph{A$_1$SC}, \emph{A$_1$SD}, \emph{A$_2$SD} and \emph{A$_2$SDF}.
The results show:
(1) Using better networks (ResNet-101 vs. VGG16) can moderately improve the performance.
However, even with state-of-the-art network \emph{A$_2$},
visual relationship detection could not be done effectively using appearance information alone.
(2) The combination of appearance and spatial configs considerably outperforms each component alone,
suggesting that visual appearances and spatial configurations are complementary to each other.
(3) The statistical dependencies are important. However, CRF is not able to effectively exploit them.
With the use of DR-Net, the performance gains are significant.
We evaluated the perplexities of the predictions for our model \emph{with} and \emph{without} DR-Net,
which are $2.64$ and $3.08$. These results show the benefit of exploiting statistical dependencies for joint recognition.

Table~\ref{tab:itr_sample} further shows the predicted relationships on several example images.
The first two columns show that the incorporation of spatial configuration can
help detect positional relationships.
The third column shows that the use of statistical dependencies can help
to resolve the ambiguities in the relationship predicates.
Finally, the fourth column shows that for subtle cases,
DR-Net can identify the relationship predicate more accurately than
the config that relies on CRF.

\begin{figure}
    \centering
    \includegraphics[width=0.4\textwidth]{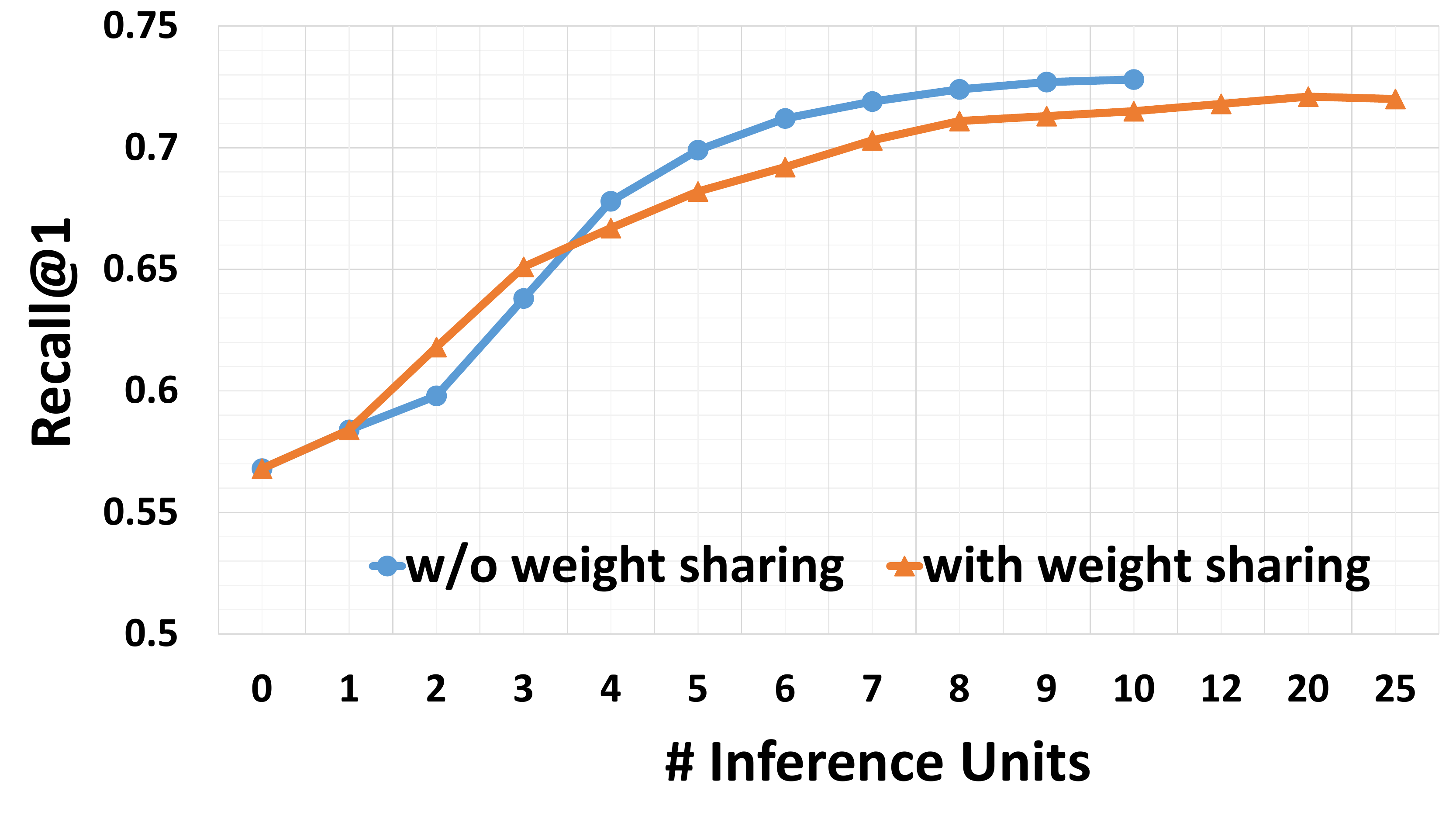}
    \caption{\small
	This figure shows the recall curves of two possible settings in DR-Net.
        In each setting, we change the number of inference units to see how the recall changes.}
    \label{fig:drnet_curve}
\end{figure}

\paragraph{Compare architectural choices.}

This study is to compare the effect of different choices in the DR-Net architecture.
The choices we study here include:
the number of inference units and whether the relational weights are shared
across these units.
The comparison is conducted on \emph{sVG}.

Figure~\ref{fig:drnet_curve} shows the resultant curves.
From the results we can see:
(1) On both settings, the recall increases as the number of inference units increases.
The best model can improve the recall from $56\%$ to $73\%$, as the number
of inference units increases.
With weight sharing, the recall saturates with $12$ inference units;
while without sharing, the recall increases more rapidly, and saturates when it has $8$ inference units.
(2) Generally, with same number of inference units,
the network without weight sharing performs relatively better,
due to the greater expressive power.


\begin{table}
    \centering
   \small
    \begin{tabular}{c|c|c|c|c}
	\toprule
	\multicolumn{5}{c}{Average Similarity} \\
	\midrule
        VR \cite{lu2016visual} & A$_1$ & S & A$_1$S & A$_1$SD \\
	\hline
	    0.2076 & 0.2081 & 0.2114 & 0.2170 & \textbf{0.2271} \\
	\bottomrule
    \end{tabular}
    \caption{\small This table lists the average similarities between generated scene graphs and the ground truth.
	All methods are named after their visual relationship detectors.}
    \label{tab:sg_rst}
\end{table}

\begin{figure}
    \centering
    \begin{tabular}{c}
    \toprule
    \includegraphics[width=0.42\textwidth]{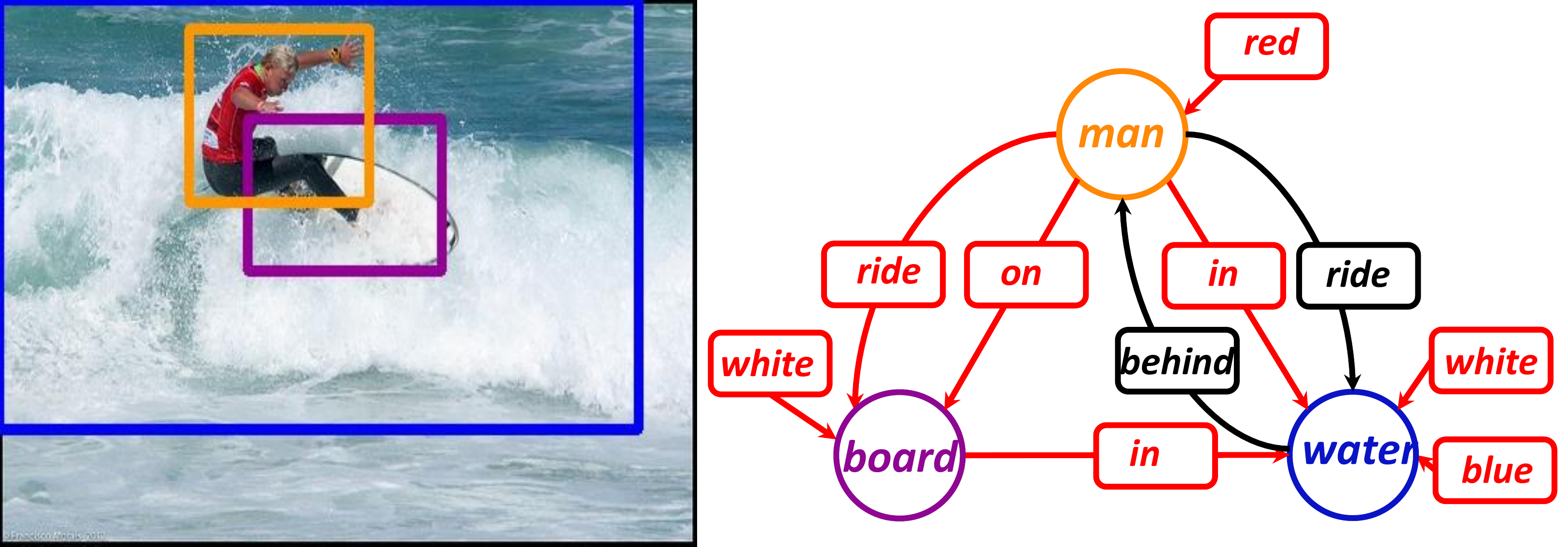} \\
    \hline
    \includegraphics[width=0.42\textwidth]{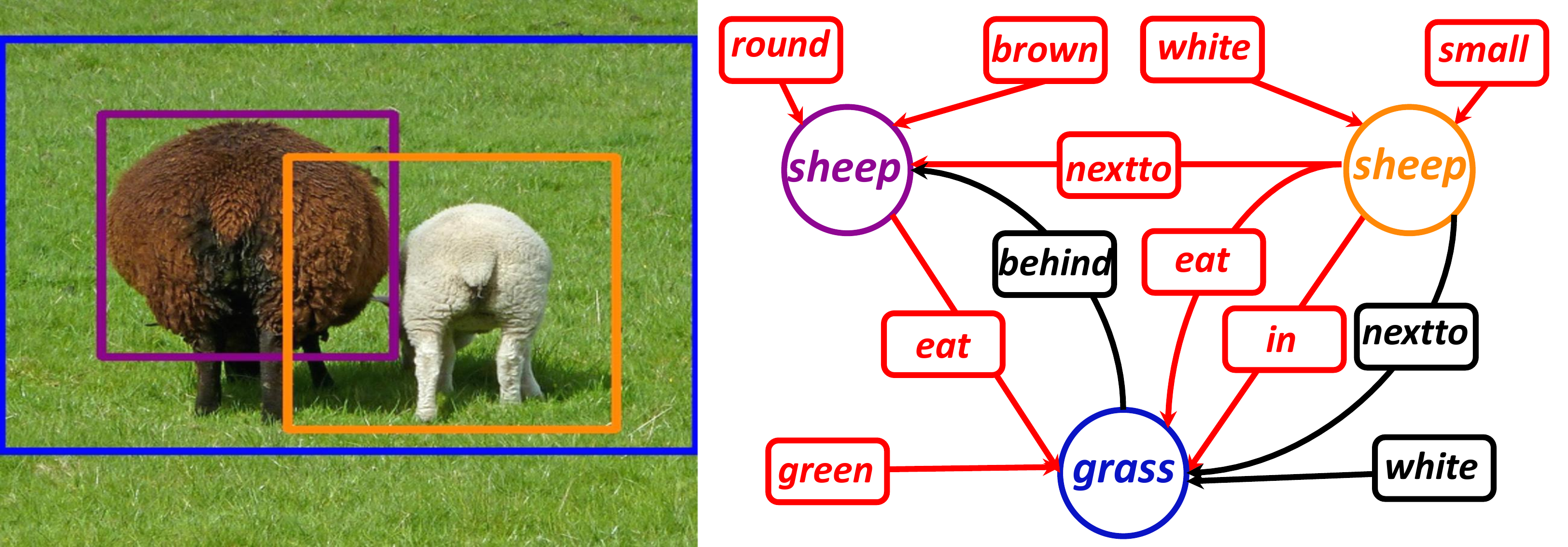} \\
    \hline
    \includegraphics[width=0.42\textwidth]{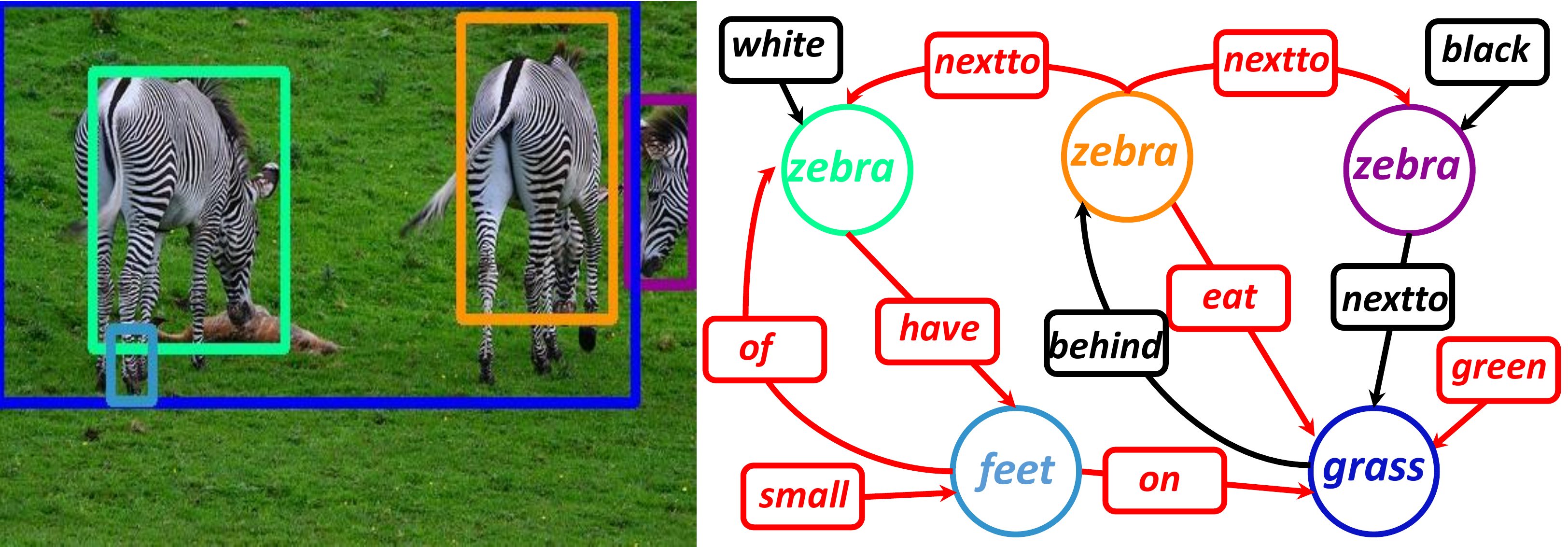} \\
    \bottomrule
    \end{tabular}
    \caption{\small This figure illustrates some images and their corresponding scene graphs.
	The scene graphs are generated according to section \ref{sec:sg}.
        In the scene graphs, the \textbf{black} edges indicate wrong prediction,
        and the \textcolor{red}{red} edges indicate correct prediction.}
    \label{fig:sg_sample}
    \vspace{-1.5mm}
\end{figure}

\subsection{Scene Graph Generation}
\label{sec:sg}

Our model for visual relationship detection can be used for scene graph generation,
which can serve as the basis for many tasks,
\eg~image captioning\cite{anderson2016spice, aditya2015images},
visual question answering\cite{wu2016visual} and image retrieval\cite{johnson2015image}.

The task here is to generate a directed graph for each image
that captures objects, object attributes, and the relationships between them~\cite{johnson2015image}.
See Figure~\ref{fig:sg_sample} for an illustration.
We compared several configs of our method, including \emph{A$_1$}, \emph{S}, \emph{A$_1$S} and \emph{A$_1$SD},
with \emph{VR}~\cite{lu2016visual} on this task,
on a dataset \emph{sVG-a}, which extends \emph{sVG} with attribute annotations.
All methods are augmented with an attribute recognizer.

For each test image, we measure the similarity~\cite{champin2003measuring} between the generated scene graph and the ground truth.
We report average similarity over all test images as our metric.
Table \ref{tab:sg_rst} compares the results of these approaches,
where \emph{A$_1$SD} achieves the best result.
This comparison indicates that
with better relationship detection, one can obtain better scene graphs.

\section{Conclusion}
\label{sec:concls}

This paper presented a new framework for visual relationship detection,
which integrates a variety of cues: appearance, spatial configurations,
as well as the statistical relations between objects and relationship predicates.
At the heart of this framework is the \emph{Deep Relational Network (DR-Net)},
a novel formulation that extends the expressive power of deep neural networks
to relational modeling.
On Visual Genome, the proposed method not only outperforms the state of the art
by a remarkable margin,
but also yields promising results in scene graph generation, a task that
represents higher level of image understanding.
These experimental results clearly demonstrate the significance of
statistical relations in visual relationship detection,
and DR-Net's strong capability in modeling complex relations.

\vspace{-10pt}
\paragraph{Acknowledgment}

This work is partially supported by
the Early Career Scheme (ECS) grant (No. 24204215), and
the Big Data Collaboration grant from SenseTime Group.

{\small
\bibliographystyle{unsrt}
\bibliography{visrel}
}

\end{document}